\documentclass[sigconf,natbib=false]{acmart}

\usepackage{algorithm,algorithmicx,algpseudocode}
\usepackage{booktabs}
\usepackage{verbatim}

\AtBeginDocument{%
  }

\setcopyright{acmcopyright}
\copyrightyear{2023}
\acmYear{2023}
\acmDOI{XXXXXXX.XXXXXXX}

\acmConference[WWW '24]{The 2024 ACM Web Conference}{MAY 13 - 17, 2024}{Singapore}

\RequirePackage[
  datamodel=acmdatamodel,
  style=acmnumeric,
  ]{biblatex}

\DeclareUnicodeCharacter{0301}{\'{e}}

\begin{document}

\title{SSTKG: Simple Spatio-Temporal Knowledge Graph for Intepretable and Versatile Dynamic Information Embedding}

\author{Ruiyi Yang}
\email{ruiyi.yang@student.unsw.edu.au}
\affiliation{%
  \institution{University of New South Wales}
  \city{Sydney}
  \state{NSW}
  \country{Australia}
}

\author{Flora D. Salim}
\email{flora.salim@unsw.edu.au}
\affiliation{%
  \institution{University of New South Wales}
  \city{Sydney}
  \state{NSW}
  \country{Australia}
}

\author{Hao Xue}
\email{hao.xue1@unsw.edu.au}
\affiliation{%
  \institution{University of New South Wales}
  \city{Sydney}
  \state{NSW}
  \country{Australia}
}

\renewcommand{\shortauthors}{Ruiyi et al.}

\begin{abstract}
Knowledge graphs (KGs) have been increasingly employed for link prediction and recommendation using real-world datasets. However, the majority of current methods rely on static data, neglecting the dynamic nature and the hidden spatio-temporal attributes of real-world scenarios. This often results in suboptimal predictions and recommendations. Although there are effective spatio-temporal inference methods, they face challenges such as scalability with large datasets and inadequate semantic understanding, which impede their performance. 
To address these limitations, this paper introduces a novel framework - Simple Spatio-Temporal Knowledge Graph (SSTKG), for constructing and exploring spatio-temporal KGs. To integrate spatial and temporal data into KGs, our framework exploited through a new 3-step embedding method. Output embeddings can be used for future temporal sequence prediction and spatial information recommendation, providing valuable insights for various applications such as retail sales forecasting and traffic volume prediction. Our framework offers a simple but comprehensive way to understand the underlying patterns and trends in dynamic KG, thereby enhancing the accuracy of predictions and the relevance of recommendations. This work paves the way for more effective utilization of spatio-temporal data in KGs, with potential impacts across a wide range of sectors.
\end{abstract}

\keywords{Knowledge graph, Spatio-temporal data, Time series forecasting}

\maketitle

\section{Introduction} \label{intro-1}

Knowledge graphs (KGs) are directed graphs comprising entities (nodes), their attributes, and the relationships between them. They represent information as facts using a node-edge-node structure. For instance, the triplet (Macdonald-compete-Burger King) represents a competitive relationship between Macdonald and Burger King. KGs adeptly capture intricate relationships between entities, enabling more contextually rich and accurate predictions. By encoding millions of real-world events or facts into graphs, KGs facilitate various downstream tasks such as recommendation system \cite{wang2019knowledge}, information retrieval \cite{dietz2018utilizing}, and question answering \cite{huang2019knowledge}. \textit{Knowledge graph completion} (KGC) methods assist KG construction by inferring missing facts based on existing ones in KGs. They learn the embedding of entities and relations on known facts and apply score functions on all possible facts to compute the possibility the fact exists, like transE \cite{bordes2013translating}, to help enhance the comprehensiveness and utility of the KG. 

\begin{figure}[!tbp]   
\centering         
\includegraphics[width=0.38\textwidth]{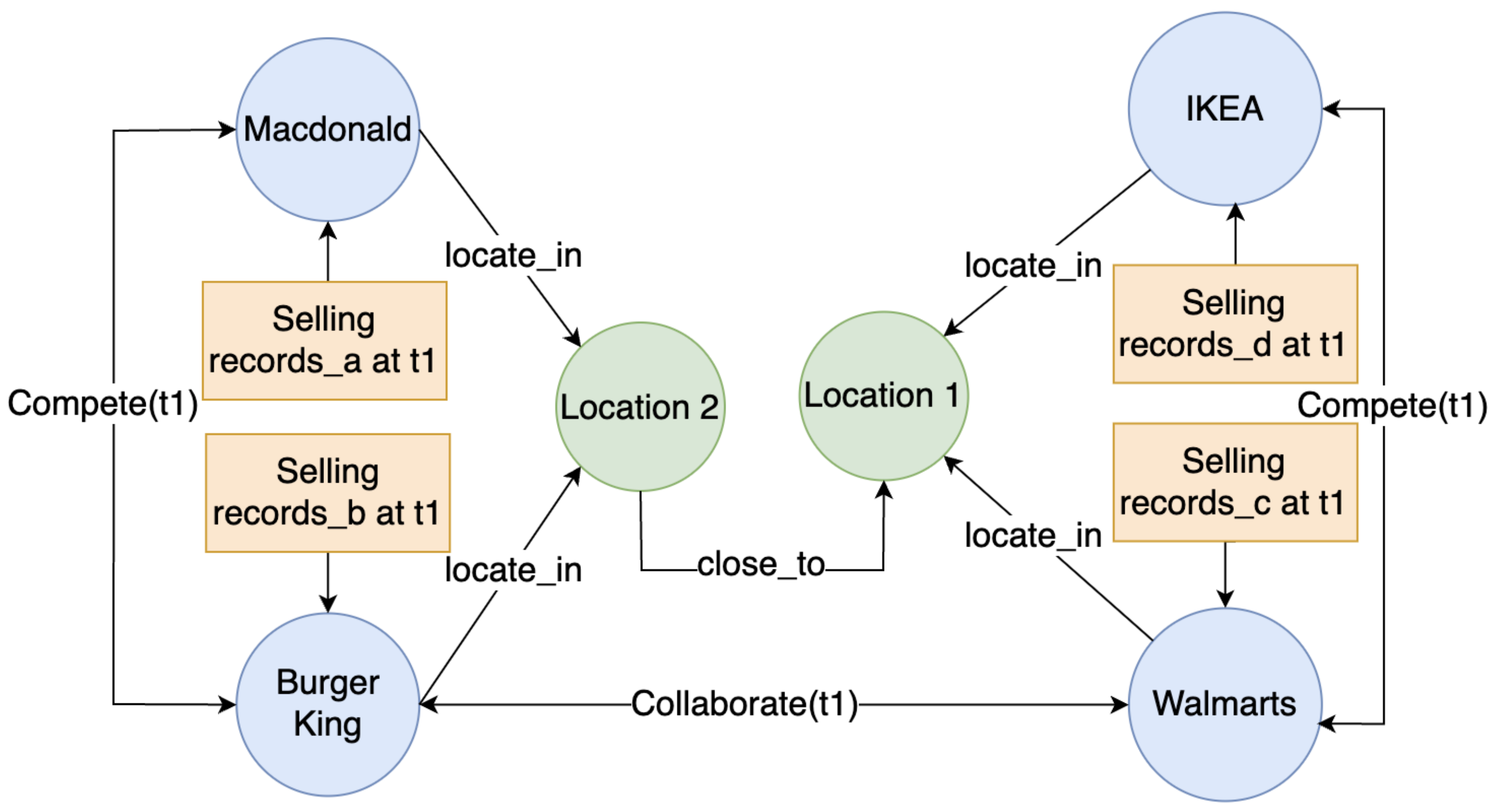}
\caption{An example of spatio-temporal KG}
\label{STKG example}
\vspace{-4ex}
\end{figure}

In practical scenarios, historical facts influence potential future relations, such as retail sales and traffic. spatio-temporal data, inherently \textit{dynamic and complex}, exhibits dependencies and relationships that evolve across time and space. The dynamic features of the data complicate the construction and maintenance of KGs that represent data comprehensively and factor in geographical relationships between entities. Static KGC methods treat facts as time-independent, leading to relation and entity embeddings stagnant, which is unrealistic \cite{chen2020knowledge}.  Many methods are raised towards temporal KG construction and completion \cite{lin2020tensor, bai2021tpmod, goel2020diachronic, messner2022temporal}. Without using KGs, a myriad of spatio-temporal prediction and recommendation methods have been proposed, yielding promising outcomes across various tasks \cite{wang2023towards}. Traditional statistical and machine learning methods like ARIMA \cite{williams2003modeling}, have been complemented by more recent deep learning methods, notably graph convolutional networks \cite{zhu2022kst}. 

Although the above methods show effectiveness when dealing with dynamic data, they still harbor notable limitations. Deep learning methods also struggle when capturing the intricate, non-linear relationships endemic to spatio-temporal data, and may fall short of incorporating broader contextual information. Data sparsity posed another challenge, constraining the improvement of their recommendation performance \cite{chen2022building}. KGs can alleviate the issues occurred using DL methods, courtesy of their rich semantics information. 

For clarity, Fig \ref{STKG example} showcases an STKG tailored for the physical store sales within a city. While various retail outlets like Walmarts, IKEA, are depicted alongside entities from different sectors, such as McDonald. Their sales records, inherently temporal, are typically tabulated over a time interval $\Delta T$, including records in several time intervals $\Delta t_{1,2,...}$. In a static KG, entities might be linked through relations like \textit{competing}, \textit{collaborating}, etc. However, in a temporal KG, these relations might evolve over time based on sales data or other practical considerations. Beyond temporal aspects, entities also exhibit geographical relations, heavily influenced by \textit{locations} and \textit{distances} separating them. Entities within STKGs can be expressed using triplets, $(Shop, t_i, loc)$, signifying an entity’s state at a specific time. Meanwhile, relations are represented as $(e_i, relation, e_j)$ under a given ${time, location(i,j)}$ highlighting the spatio-temporal connection between two entities. By integrating these triplets, a semantic path is constructed, elucidating the evolution of relationships grounded in spatio-temporal data.

STKGs are versatile tools suitable for variety of predictive and recommendation tasks. However, training a spatio-temporal KG on benchmark datasets like Wikidata \cite{leblay2018deriving, lacroix2020tensor} or YAGO15K \cite{garcia2018learning} proves time-consuming \cite{cai2022temporal}. \textit{The time cost is magnified when applied to extensive real-world datasets.} The dynamic nature of data and intricate relationships within the graph present challenges in harnessing an STKG effectively for downstream applications, which leads to research questions: \textbf{Is there a simple and effective way to construct and exploit an STKG versatile enough to accommodate diverse data types? How to enable this framework for KG completions while ensuring its interpretability?}

In this paper, a novel framework - Simple Spatio-Temporal Knowledge Graph (SSTKG) is raised for constructing and exploring spatio-temporal knowledge graphs for prediction and recommendation. By integrating spatio-temporal data into KGs and exploiting these KGs through entity and relation embeddings, the framework aims to leverage the strengths of KGs to enhance the accuracy and relevance of spatio-temporal predictions, while ensuring efficiency as well as interpretability. To validate its efficacy, the framework was applied to two datasets:  Safegraph's Spend-Ohio dataset and the Traffic Volume of New South Wales (TFNSW) dataset to do experiments on temporal sequence prediction.

\section{Related Work}

\subsection{Spatio-Temporal Data Prediction}

In the early stages, methods are based on statistical knowledge, or using machine learning, such as Autoregressive Integrated Moving Average (ARIMA) \cite{williams2003modeling}, Support Vector Regression (SVR)~\cite{hearst1998support}, Critical Support Vector Machine (CSVM)~\cite{raicharoen2003application} These methods focus more on numerical data. Although they consider number series to be related to time, they are unable to capture enough spatio-temporal dependencies. More recently, researchers have begun to consider more complex methods, thanks to the emergence of deep learning. Long Short-Term Memory (LSTM) networks \cite{gers2002learning}: It's a type of Recurrent Neural Networks(RNNs) outperforming other RNNs particularly in learning long-term dependencies. Gated Recurrent Unit (GRU) RNNs \cite{cho2014properties} can capture dependencies of different time scales, controling the information flow from the previous activation when computing the new. \cite{chung2014empirical}. Temporal Convolutional Networks(TCNs) \cite{bai2018empirical} can effectively capture temporal features with an architecture of sequence modeling, causal convolutions, dilated convolutions, and residual connections. Based on TCN, Multivariable Temporal Convolutional Networks (M-TCN) \cite{wan2019multivariate} allows a model for multi-variable time series prediction.

The above models are efficient in handling temporal patterns inside sequences, however, sensors collecting data may also be spatially related, which is omitted by the above models. Graph Neural Networks(GNNs)~\cite{zhou2020graph} are neural models that capture the dependence of graphs via message passing between nodes of graphs. Based on that, Temporal Graph Convolutional Network (T-GCN) \cite{zhao2019t} is combined with the GCN and the GRU. By using GCN to learn topological structures to learn spatial dependence and using GRU to learn temporal patterns. Moreover, Spatio-Temporal Graph Convolutional Networks (ST-GCN) \cite{yu2017spatio} utilizes graph CNNs for extracting spatial features, and gated CNNs for extracting temporal features. A spatio-temporal convolutional block is used to fuse the above two patterns. The fusion of deep learning models leads to a great fit for spatio-temporal data. spatio-temporal GNNs can simultaneously model spatial and temporal information \cite{defferrard2016convolutional}, and dealing with real-world related works like Traffic flow prediction \cite{diehl2019graph, xie2020sast, wang2020traffic}, Next POI recommendation \cite{han2020stgcn}, Crime prediction \cite{han2020risk}, Weather forecasting \cite{keisler2022forecasting} and Human action recognition \cite{wang2022skeleton}.

\subsection{Knowledge Graph for prediction}

\subsubsection{Static KGs for Prediction}

Since KGs have a unified structure, based on their embeddings or paths, they can be used to predict potential links hidden in established datasets. For static data, KGs can assist and accelerate drug discovery \cite{zeng2022toward} in the medical field, and they also perform well on fake news detection \cite{ciampaglia2015computational} by finding the shortest path between facts.

\subsubsection{Temporal KGs/STKGs for Prediction}

Dynamic data, typically sourced from sensors, can also be transferred into structured entities and shaped into temporal KGs(TKGs) or STKGs. Embeddings encompassing distinct spatial or temporal information are compared to determine the entities that would appear in certain time points under certain locations. Since dynamic KGs capture time relationships between entities in events, temporal predictions like the time of natural disasters \cite{ge2022disaster} could be achieved. STKGs also help in spatial predictions, by modeling trajectories data,  users' mobility patterns or activities can be predicted \cite{wang2021spatio, chen2022building}.

\subsection{Knowledge Graph for recommendation}

Apart from prediction, knowledge graphs are also widely used for recommendation. While prediction works (sequence prediction, event prediction, POI prediction, etc.) are highly related to the dynamic nature of entities evolution over time, large proportions of recommendation systems on KG are related to the structure of KG (like entity properties or relation properties), which are related to graph algorithms (path searching) or embedding techniques.

\subsubsection{Path-based recommendation} Paths in KGs contain relationships between entities, enabling the extraction of features such as users' preferences or item characteristics by analyzing paths. KPRN \cite{wang2019explainable} used the LSTM network to represent path information, like users and movie interactions, thus can calculate user preferences towards target movies.

\subsubsection{Embedding-based recommendation} Entities and relations can normally be transferred into embeddings under certain rules, these embeddings can be applied to recommendation algorithms. Entity2rec \cite{palumbo2020entity2rec} uses property-specific embeddings on KGs to do recommendation, while HAKG \cite{sha2021hierarchical} uses subgraph embeddings for enhanced user preference prediction.

While the aforementioned methodologies have registered good performances in designated tasks, they are encumbered by certain limitations: 1) Their inherent complexity or the extensive versatility of entities often renders them time-intensive or restricts adaptability to diverse domains. 2) They are not explainable enough to describe features extracted from spatio-temporal data. The proposed model aspires to bridge these gaps, presenting streamlined a solution, adaptable to diverse data types while ensuring interpretability.

\section{Methodology}

\subsection{Preliminaries and method overview}

The STKG problem is defined as: An optimal STKG should accommodate the dynamic nature of data, adapting to changes in entities' attributes influenced by time and location, facilitating the completion and enhancement of KG after construction, and predicting forthcoming attributes and relationships for entities.
Table~\ref{notation} summarizes notations used in the paper as well as their meanings.

\textbf{Input representation} The objective of the proposed STKG is to attain universality. To this end, a uniform representation for diverse types of spatio-temporal data is integrated to generalize raw entities and relations types.

\textbf{STKG embedding model} The embedding model is designed to encode entity attributes into vector representations and subsequently decode embeddings into numerical representations mirroring the raw data. The embedding model facilitates KG completion on existing STKG and enables the prediction of underlying or between entities.

\begin{table} [!tbp]
\caption{Notations and descriptions}
  \begin{tabular}{ccl}
    \toprule
    Notation & Description\\
    \midrule
    e, r & An entity and a relation \\
    \textbf{e}, \textbf{r} & Vector representation of e and r \\
    $r_{i,j}$ & directional relation from i to j \\
    \emph{E}, \emph{R} & Entity set and relation set\\
    $d(e_i, e_j)$ & Distance between two entities \\
    \emph{G} & A STKG \\
    \emph{T} & The set of time \\
    \textbf{$e_{attribute}$} & The embeddings of certain attribute of entities \\
    $I_{(e_i,e_0)}$ & Influence that $e_i$ applied on $e_0$ \\
    $W_{(e_i,e_0)}$ & Weight variable used during training influence \\
  \bottomrule
\end{tabular}
\label{notation}
\end{table}

\begin{figure}[htbp]   
\centering         
\includegraphics[scale=0.25]{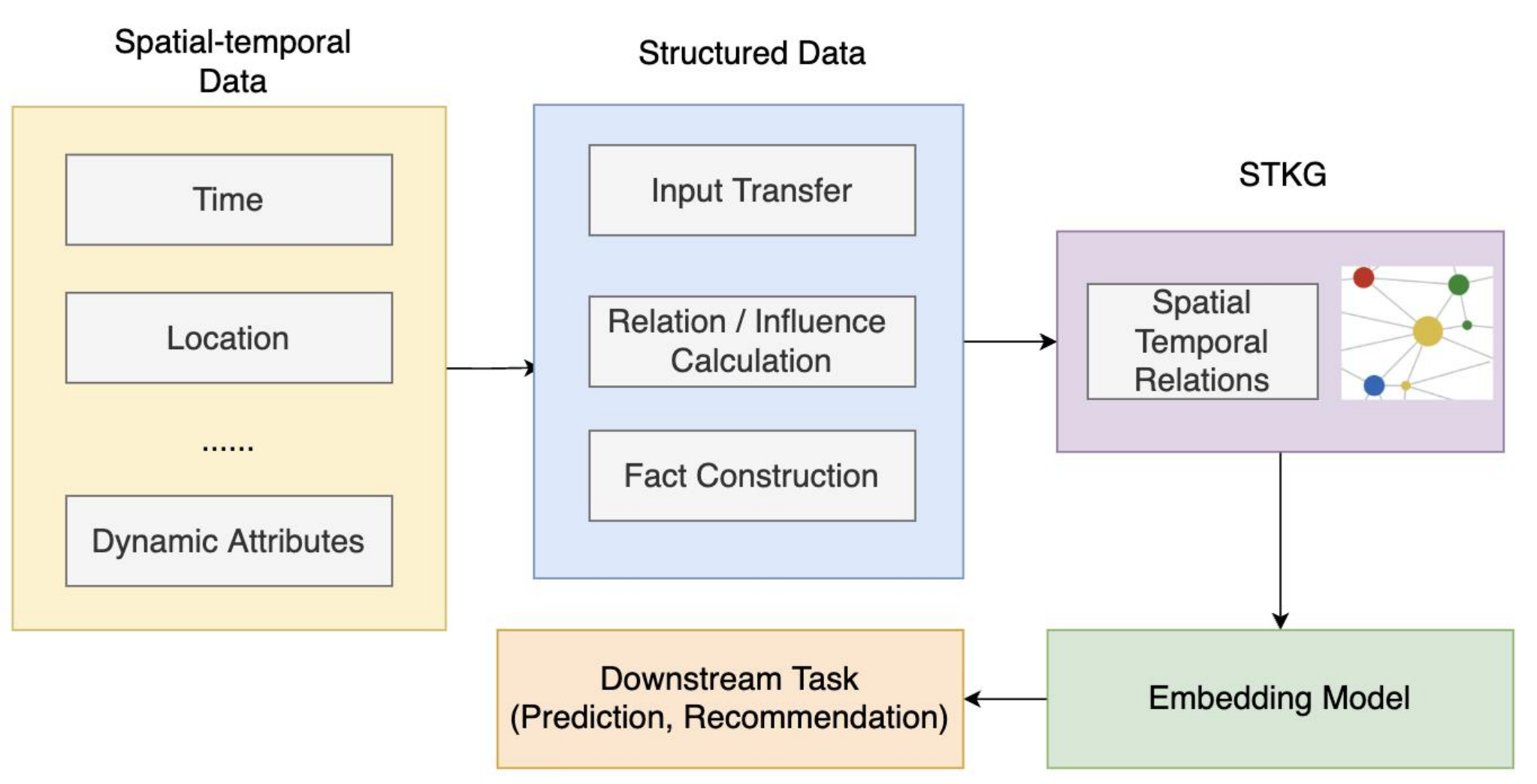}
\caption{The workflow of proposed framework}
\label{workflow}
\end{figure}

Figure~\ref{workflow} illustrates the general workflow of the framework. Upon acquiring spatio-temporal data, pre-established rules are utilized to extract and compute entities, relationships, and facts, thereby constructing STKG while \textit{ensuring limited entity types and relationship types}. Subsequently, a new embedding model is raised to vectorize the features of entities, enabling the utilization of the STKG in downstream tasks. This streamlined process facilitates a more efficient and effective application of knowledge graphs in real-world scenarios, able to be used for inference with enhanced speed. Finally, the underlying patterns and insights captured by the STKG are interpreted based on its structure, making the whole model explainable.

\subsection{Knowledge Graph Construction} \label{method_KGC}

\subsubsection{Definition of STKG}\label{d-STKG} The spatio-temporal knowledge graph is defined as graphs \emph{G} = (\emph{E}, \emph{R}, \emph{T}, \emph{F}), where \emph{E}, as table~\ref{notation} shows, is entities that contains spatio-temporal attributes. \emph{R} represent the set of relation between entities. \emph{T} describes how the temporal records get divided. \emph{F} is the set of facts mentioned in section \ref{intro-1}.  Specifically, \emph{R} and \emph{T} in  the knowledge graph define \textit{certain relation between entities under certain time}, which denotes facts. Facts under STKG are seen as a quadruple ($e_i,t,t_j,r$).

\subsubsection{Simple STKG (SSTKG)}\label{d-SSTKG} Considering when entities like stores are rigidly classified according to their business establishments, as exemplified by the 6-digit North American Industry Classification System (NAICS) code. The strict categorization can lead to an excessive fragmentation of entity types. Also, the dynamics of relationships between entities can vary significantly based on spatial and temporal factors. Two entities, even if their spatial distances are fixed, might have totally inverted relations at different times. Moreover, detailed numerical time and location are hard to be transferred as distinct entities.

In light of these complexities, the simple STKG (SSTKG) aims to provide a more flexible and realistic representation of entities and their relationships, establishing rules for the SSTKG as follows:
\begin{itemize}
    \item \emph{Rule 1:} Time and location are not treated as independent entities. Instead, they are integrated as attributes inherent and between entities, represented as part of entity and relation embeddings. 
    \item \emph{Rule 2.} The model prioritizes a reduction in the number of entity types, embedding classification data directly within the entity. This not only simplifies the graph structure but also facilitates more efficient and direct retrievals of classification information from the entity embeddings.
    \item \emph{Rule 3.} Numerical representations are adopted to directly articulate the relationship between two entities. Under this paradigm, the association between entities is conceptualized as a continuous variable termed ``influence''. Within this framework, any pair of entities can exhibit a relationship that is fluid across both temporal and spatial dimensions
    \item \emph{Rule 4:} Relationships between entities that are quantitatively negligible are omitted, ensuring focus on significant interactions and reducing noise within the graph.
\end{itemize}

Leveraging this SSTKG framework, entities are directly extracted from structured data. The process of relation extraction is thus transformed into ``relation computation'', or ``influence computation'', while fact still be seen as the quadruple ($e_i,t,t_j,r$).

\subsubsection{Algorithm for constructing SSTKG}

The detailed process of constructing the SSTKG is elucidated according to \ref{d-SSTKG}. The temporal records for an entity are viewed as \textit{The result of related entities applying influence plus itself's basic record}, which is:

\begin{equation} \label{eq0}
 p_{e_0}*Record_{e_0}(t) = \Sigma_{i=1}^n I_{(e_i,e_0)}Record_{e_i}(t)
\end{equation}

Fitting Equation~\eqref{eq0} is seen as a regression process, where 1-p is seen as a parameter quantifying the self-influence of an entity, providing a measure of how much an entity's characteristics contribute to its own behavior or status within the knowledge graph. While temporal variable t represents a time slot, the integration of temporal data and spatial relationships facilitates the computation of a relation ``weight''~\eqref{eq1}, using overall record and distance between two entities, is seen as a ratio of properties of $e_i$ to $e_0$. The p in~\eqref{eq0} is counted in Equation~\eqref{eq2}. Then the influence that entity $e_i$ may apply on $e$ during time slot t is calculated in Equation~\eqref{eq3}:

\begin{equation} \label{eq1}
  W_{(e_i,e_0)} = \frac{OverallRecord(e_i)}{OverallRecord(e_0)}  * log(1+\frac{\Sigma_{j=i}^n Distance(e_j,e_0)}{n*Distance(e_i,e_0)})
\end{equation}

\begin{equation} \label{eq2}
  p_{e_0} = \frac{\Sigma_{i} W_{(e_i,e_0)}}{\Sigma_{k,j} W_{(e_k,e_j)}}
\end{equation}

\begin{equation} \label{eq3}
  I_{(e_i,e_0)} = regressionFactor*W_{(e_i,e_0)}
\end{equation}

Algorithm \ref{alg:construct} shows the pseudocode for constructing SSTKG, yt needs to be emphasized that, the ``influence'' is unidirectional. In determining the ``influence'', only the spatio-temporal information of entities is considered. Attributes of entities, such as categories, remain unaddressed. Such an omission in SSTKG construction arises from potential complexities in the data; for instance, the prevalence of numerous categories as seen with the NAICS code shown in Section \ref{method_KGC}. On the other hand, some data is hard to fit entities in specific categories, like traffic volume data. Hence, these data are integrated into KG embedding, as elaborated in Section \ref{emb}.

\begin{algorithm}
\caption{Constructing a SSTKG using time-series records data}\label{alg:construct}
\begin{algorithmic}[1]
\Require Entity \(E\), Location \(L\), time-series records \(TS\), distance threshold \(D\)
\Ensure Quadratic relation set \(R\)
\For{\(e \in E\)}
    \State filtering \(E_0 \subseteq E\) where 
    \ForAll{\(e_i \in E_0\)}
        \If{\text{Distance}(\(e,e_i\)) \(\leq D\)}
        \EndIf
    \EndFor

    \For{\(e_i \in E_0\)}
        \State \(W_{(e_i,e)}\) \(\gets\) Compute weight using (\eqref{eq1})
    \EndFor
    \State \(p_e\) \(\gets\) Compute \(p\) using (\eqref{eq2})
    \State \(influence_{(E_0,e)}\) \(\gets\) Compute influence using (\eqref{eq3})
\EndFor
\end{algorithmic}
\end{algorithm}

\subsection{Embedding Model} \label{emb}

One entity's temporal data record as well as its spatial location is assumed to influence other entities' temporal records. While the numerical ``influence'' is seen as a relation, the embedding model aims to map attributes of entities and relations into low-dimensional vectors. Embeddings generated by the model are further implemented into downstream work. Specifically, the embeddings are categorized into 3 boxes: static, dynamic in and dynamic out.

\subsubsection{Static Embedding}
This component encapsulates the static attributes of an entity, yielding a representation that remains invariant over time. Static attributes are left when calculating ``influence''. However, in the computation of the static embedding, these attributes that were previously set aside are reintegrated. Apart from categorical attributes, a summary of the entity's comprehensive spatio-temporal data is integrated into the static embedding. Metrics such as average sales volume or average traffic flow are included to represent the ``magnitude'' or ``scale'' of the entity. Equation~\eqref{eq4} shows the formation of static embedding, where $\phi$ manages to regularize overall records into a smaller range. 

\begin{equation} \label{eq4}
  e_{i\_static} = e_{i\_category}* \phi(overall\_records)
\end{equation}

\subsubsection{Dynamic Embedding} Dynamic embedding contains directions of entity relationships. With the use of spatial-temporal records, it is formed by two subsets: out embedding and in embedding, reflecting ``pointing to'' and ``pointing from'' links between entities on a knowledge graph.

Out embedding signifies the potential influence an entity may impart upon its linked entities. It is configured as the dynamic embedding representing the ``influence level'' of the entity itself, disregarding spatial relationships with other entities. The computation of the out embedding is shown in Equation~\eqref{eq5}, encompassing concatenation of the static embedding with its temporal records. The out embedding is a combination of the entity's overall status and temporal status.

\begin{equation} \label{eq5}
  e_{i\_out}^t = \psi(e_{i\_static},e_{i\_records}^t)
\end{equation}

In Embedding quantifies the influence that an entity receives from its associated entities, reflecting the cumulative impact of these relationships on the entity. Analogously, in the formation of the SSTKG, the embedding is viewed as an aggregate of the entity's inherent influence and the influences exerted by its associated entities. Shown in Equation~\eqref{eq6}, p is the weight shown in Equation~\eqref{eq2}. On vector space it is represented in Equation~\eqref{eq6-2}.

\begin{equation} \label{eq6pre}
  p_i*e_{i\_out}^t = e_{i\_in}^t
\end{equation}

\begin{equation} \label{eq6}
  \mathbf{e_{i\_in}^t} = \Sigma_j F(Influence_{(i,j)} * \mathbf{e_{j\_out}^t}) 
\end{equation}

\begin{equation} \label{eq6-2}
  \mathbf{e_{i\_in}^t} = \Sigma_j (Influence_{(i,j)}* \mathbf{e_{j\_record}^t} + \mathbf{e_{j\_static}})
\end{equation}

\subsubsection{Embedding Training Algorithm}  

The output of Equation~\eqref{eq6} in the embedding model is not directly ascertainable, since after adding the influence, out-embedding needs to be trained to fit the equation, which leads to modification in static embedding. The static embedding and out-embeddings are used as input, optimized embeddings are obtained after training.

Let $E_0$ represent a set of out embeddings of entities that have potential relations, with entity $e_0$, while set $R$ denotes the initial influence of entities in $E_0$ as (1*n) vector to $e_0$. Given a training tuple x = ($e_0$, R, $E_0$, t) Equation~\eqref{eqscore1} defined a score as how precise one entity is \textbf{influenced} by related entities. Meanwhile, valid relations and embedding sets are used to obtain a lower score of $f_{SSTKG}$, then the first loss function is defined in Equation~\eqref{eqloss1}.

\begin{equation} \label{eqscore1}
  f_{p1}(x) = f_{1}(e_0, R, E_0, t) = ||p_{e_0}*e_{0\_out}^t - e_{0\_in}^t||_2^2
\end{equation}

\begin{equation} \label{eqloss1}
  l_{emb}(x) = l_{emb}(e_0, R, E_0, t) =  -\Sigma_i log \sigma(f_{p1}(e_0, R^i, E_0, t)-f_{p1}(x)) 
\end{equation} 

Embedding $e_{out}^i$ is replaced in $f_{p2}(e_0, R^i, E_0, t)$ with another random entity that has similar overall selling records in the whole dataset and without relations with $e_i$, using the same Influence value. An alternate score function is defined for the loss of entity influence values. For an entity $e_0$, its influence on the SSTKG, which is related to entities $E_0$ that connect with $e_0$, $R$ now denotes after optimizing out-embeddings, the influence of $e_0$ to entities in $E_0$. Given a training tuple x = ($e_0$, R, $E_0$, t), the score function is articulated as:

\begin{equation} \label{eqscore2}
  f_{p2}(x) = f_{p2}(e_0, R, E_0, t) = ||p_{e_0}* e_{0\_out}^t - \Sigma_i R_i * e_{i\_{out}}^t||_2^2
\end{equation}

\begin{equation} \label{eqloss2}
  l_{inf}(x) = l_{inf}(e_0, R, E_0, t) =  -\Sigma_i log \sigma(f_{p2}(e_0, R, E_0^i, t)-f_{p2}(x)) 
\end{equation} 

In $f_{SSTKG}(e_0, R^i, E_0, t)$ one related entity's influence is replaced to average. The second loss function denotes the loss of specific \textbf{``influence''} value, which is the relations. Algorithm \ref{alg:two} shows the process of learning improved embeddings and influences.

\begin{algorithm}
\caption{Training entity embeddings and relations for SSTKG}\label{alg:two}
\begin{algorithmic}[1]
\Require \(N_{epochInf}\), \(N_{epochEmb}\), SSTKG \(G\) with initialized \(e_{static}\), \(e_{out}\), influence
\Ensure SSTKG with trained \(e_{out}\), influence
\For{\(i = 1\) to \(N_{epochInf}\)}
    \State \(S_1 \gets G\)
    \While{\(S_1 \neq \emptyset\)}
        \State Sample batch \(S_{batch} \subset S_1\)
        \State \(S_1 \gets S_1 \setminus S_{batch}\)
        \State \(L_1 \gets 0\)
        \For{\(s \in S_{batch}\)}
            \State \(f_{p1}(s) \gets\) compute score using (\ref{eqscore1})
            \State \(l_{inf}(s) \gets\) compute loss using (\ref{eqloss1})
            \State \(L_1 \gets L_1 + l_{inf}(s)\)
        \EndFor
        \State Update out embeddings using \(\nabla L_1\)
    \EndWhile
\EndFor
\For{\(i = 1\) to \(N_{epochEmb}\)}
    \State \(S_2 \gets G\)
    \While{\(S_2 \neq \emptyset\)}
        \State Sample batch \(S_{batch} \subset S_2\)
        \State \(S_2 \gets S_2 \setminus S_{batch}\)
        \State \(L_2 \gets 0\)
        \For{\(s \in S_{batch}\)}
            \State \(f_{p2}(s) \gets\) compute score using (\ref{eqscore2})
            \State \(l_{emb}(s) \gets\) compute loss using (\ref{eqloss2})
            \State \(L_2 \gets L_2 + l_{emb}(s)\)
        \EndFor
        \State Update influence in relations using \(\nabla L_2\)
    \EndWhile
\EndFor
\end{algorithmic}
\end{algorithm}

\section{Model properties}

\subsection{Efficiency and Speed}
The proposed model is designed with computational efficiency in mind. It requires less computational resources compared to traditional models, thereby enabling faster construction of the STKG. This feature is particularly beneficial in scenarios where rapid knowledge graph construction is crucial. Here is the test result of constructing and optimizing an SSTKG using the Spend-Ohio dataset mentioned in Section \ref{dataset}, with 100 training epochs.

\subsection{Inference Patterns}

By using the embedding model in Section \ref{emb}, a certain entity's temporal record is predicted using its related entities' records. Based on Equation~\eqref{eq6}, trained static embedding of related entities and their current temporal records are used to compute the target entity's out-embedding. Therefore, final temporal records are decoded from out embedding as well as the static embedding, since for the trained embeddings, influence$\in R$ are obtained, while having related entities' records on time slot $t_1$, the out/in embeddings for $e_0$ is inferred based on Equation~\eqref{eq6} and ~\eqref{eq6-2}. Subsequently, the referred $e_{i\_{records}}^t$ is decoded in accordance with Equation~\eqref{eq5}.

\subsection{Interpretability} \label{inter}

Another significant advantage of SSTKG is its interpretability. The simple structure and the numerical representation of relationships make it easier to understand the underlying patterns and insights captured by the STKG. This interpretability enhances the model's usability, especially in applications where understanding the reasoning behind predictions is important.

Embedding directly reflects the spatio-temporal properties of each entity based on backward induction. The whole fitting and training process, to simply explain, is a process of finding proper embeddings that incorporate an entity's spatio-temporal data, such that the embedding (out-embedding), is viewed as the result of the combined effects of related entities' embeddings(out-embedding), during which the unidirectional relation between two entities serves as the parameter of fitting the whole equation. First, an expansion of the Equation~\eqref{eq6} is resented in Equation~\eqref{eq7} and then transferred as Equation~\eqref{eq8}.

\begin{equation} \label{eq7}
  p_i*e_{i\_in}^t = \Sigma_j \psi(e_{i\_static},e_{i\_records}^t) * Influence_{j,i} \\
\end{equation}

\begin{equation} \label{eq8}
  p_i*e_{i\_in}^t = \Sigma_j e_{j\_static}*\Omega (e_{i\_records}^t, Influence_{j,i}) \\
\end{equation}

Clearly, \(\Omega\) after this transformation, served as connecting parameters of out-embeddings (the temporal record) to the influence variables: it's a temporal relation of entity \(j\) to \(i\), which is further explained as \textbf{entity \(j\)'s influence to \(i\) under time \(t\)}, also it can serve as generating an embedding of temporal relation, simplified as Equation~\eqref{eq9}.

\begin{equation} \label{eq9}
  p_i \cdot e_{i}^t = \sum_j e_{j}^t \cdot influence_{j,i} = \sum_j e_{j\_static} \cdot r_{j,i}^t
\end{equation}

Thus, from the final result, training the embedding serves to refine the processes undertaken during SSTKG construction. it optimizes the whole SSTKG, forming the exact relationship using both entities' categorical, spatial, and temporal attributes.

\begin{table}
\centering
\caption{Time cost for training SSTKG on Spend-Ohio dataset}
  \begin{tabular}{ccc}
    \toprule
    entity number & time records (day) & average time(s)\\
    \midrule
    1000 & 30 & 347.7 \\
    39188 & 30 & 15942.8 \\
    41200 & 30 & 16506.1 \\
  \bottomrule
\end{tabular}
\label{tabel-time}
\end{table}

\section{Experiments}

\subsection{Datasets} \label{dataset}

\begin{figure}[htbp]   
\centering         
\includegraphics[width=0.325\textwidth]{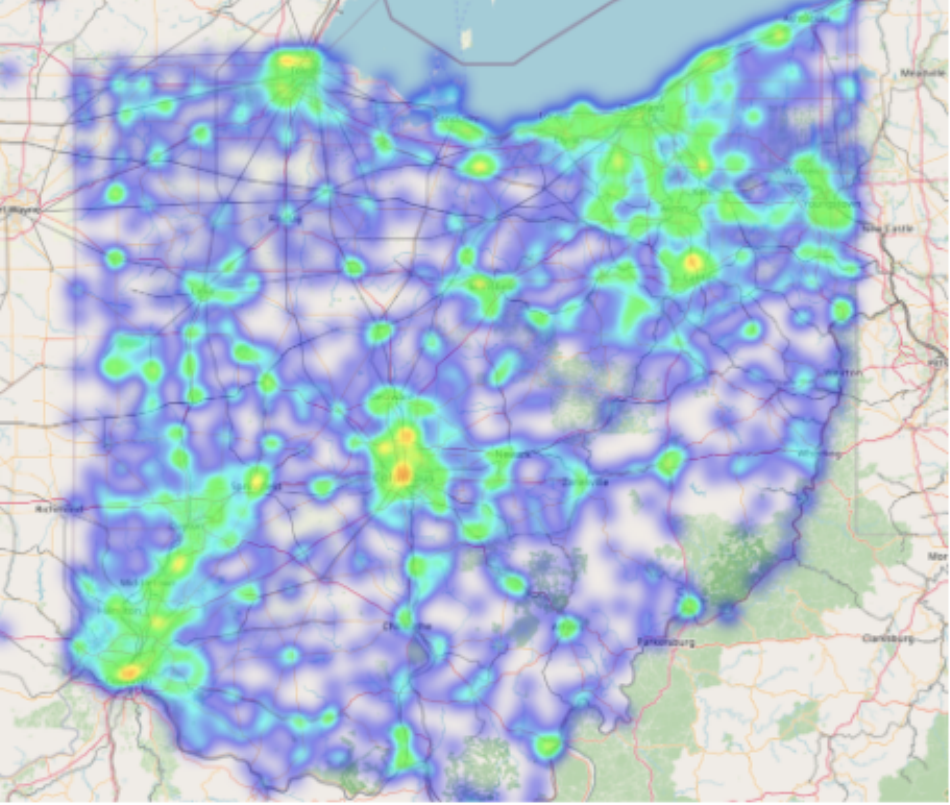}
\caption{Heatmap of stores in Spend-Ohio dataset}
\label{heatmap-Ohio}
\end{figure}

Two datasets are used to evaluate the performance of SSTKG. The first one is Spend-Ohio data from January 2022 to April 2023, collected by Safegraph,  containing many Ohio stores' geographical and categorical information, as well as the selling records \textbf{counted by day}. Figure~\ref{heatmap-Ohio} represent entities' distribution in forms of heatmap. The second one is Traffic Volume of Transport for New South Wales (TFNSW) data, which encompasses the traffic volume from a collection of permanent traffic counters and classifiers in Sydney, with data collated since 2008 on an \textbf{hourly basis}. Locations of these counters have been further categorized based on their respective suburbs. Table~\ref{quant-datasets} presents the size of the two datasets. Notably, the 'distance' attribute represents the distance threshold employed during SSTKG construction as per Algorithm \ref{alg:construct}. The attributes used in processed Spend-Ohio and TFNSW data are shown in Table~\ref{attribute-combined}.

\begin{table}[ht]
    \centering
    \caption{Quantities of data used in datasets}
    \begin{tabular}{ccccl}
        \toprule
        \multicolumn{5}{c}{\textbf{Spend-Ohio dataset}} \\
        \midrule
        data & entities & distance & relations & records\\
        \midrule
        2022-3 & 39188 & 2km & 2941374 & 1014976\\
        2022-4 & 39461 & 2km & 2970417 & 1055901\\
        2022-5 & 39654 & 2km & 3028519 & 1083649\\
        2022-6 & 39931 & 2km & 3062957 & 1098972\\
        2023-1 & 41200 & 2km & 3200018 & 1277200\\
        2023-2 & 41138 & 2km & 3194903 & 1151864\\
        2023-3 & 42932 & 2km & 3314523 & 1300893\\
        \midrule
        \multicolumn{5}{c}{\textbf{TFNSW dataset}} \\
        \midrule
        data & entities & distance & relations & records\\
        \midrule
        2015 & 67 & 4km & 496 & 1045200\\
        2016 & 69 & 4km & 511 & 1212192\\
        \bottomrule
    \end{tabular}
    \label{quant-datasets}
\end{table}

\begin{table}[ht]
    \centering
    \caption{Attributes for constructing SSTKG in datasets}
    \begin{tabular}{cp{2in}}
        \toprule
        \multicolumn{2}{c}{\textbf{Spend-Ohio dataset}} \\
        \midrule
        attribute & detail explanation  \\
        \midrule
        Placekey & a tuple representing entity location  \\
        NAICE code & 6-digit code reflecting category \\
        Temporal records & selling records collected day by day  \\
        Overall records & overall records calculated by past results  \\
        \midrule
        \multicolumn{2}{c}{\textbf{TFNSW dataset}} \\
        \midrule
        attribute & detail explanation  \\
        \midrule
        Location & counters' locations  \\
        Suburb & the suburb where counters are located   \\
        Temporal records & traffic volume collected by hour  \\
        Overall records & aggregated traffic volume   \\
        \bottomrule
    \end{tabular}
    \label{attribute-combined}
\end{table}

\begin{table*}[htbp]
    \centering
    \begin{minipage}{0.49\textwidth}
        \centering
        \caption{Test results for Spend-Ohio datasets}
        \label{combined_spendOhio}
        
        \begin{tabular}{lcccc}
            \toprule
            & \multicolumn{4}{c}{Spend-Ohio dataset: 2022.3 - 2022.6} \\
            \cline{2-5}
            method & acc@10 & acc@15 & RMS & RSD \\
            \midrule
            SVR & 0.5621& 0.6528& 0.09872 & 158.9 \\
            LSTM & 0.5984& 0.7025& 0.09031 & 135.7 \\
            GRU & 0.7057& 0.8544& 0.0607 & 97.3 \\
            T-GCN & 0.7489& 0.8386& 0.0651 & 103.5 \\
            ST-GCN & 0.7902& \textbf{0.8945}& 0.0463 & 87.9 \\
            SSTKG & \textbf{0.8016}& 0.8922& \textbf{0.0452} & \textbf{86.1} \\
        \end{tabular}
        \begin{tabular}{lcccc}
            \toprule
            & \multicolumn{4}{c}{Spend-Ohio dataset: 2023.1 - 2023.3} \\
            \cline{2-5}
            method & acc@10 & acc@15 & RMS & RSD \\
            \midrule
            SVR & 0.6015& 0.7325& 0.09751 & 144.3 \\
            LSTM & 0.6394 & 0.7672 & 0.08865 & 127.2 \\
            GRU & 0.7359 & 0.8897& 0.0528& 88.3 \\
            T-GCN & 0.7826 & 0.8597& 0.0562& 91.3 \\
            ST-GCN & \textbf{0.8435} & 0.09291 &0.0399 & 76.8 \\
            SSTKG & 0.8374 & \textbf{0.9289} & \textbf{0.0396} & \textbf{71.7} \\
            \bottomrule
        \end{tabular}
    \end{minipage}
    \hfill
    \begin{minipage}{0.49\textwidth}
        \centering
        \caption{Test results for TFNSW datasets}
        \label{combined_TFNSW}
        
        \begin{tabular}{lcccc}
            \toprule
            & \multicolumn{4}{c}{TFNSW dataset: hourly prediction} \\
            \cline{2-5}
            method & acc@10 & acc@15 & RMS & RSD \\
            \midrule
            SVR &  0.701 &0.7583 &0.06737 & 129.8 \\
            LSTM & 0.7639 & 0.8072 & 0.05615 & 113.4 \\
            GRU & 0.7825 &0.8404 & 0.0475& 107.8 \\
            T-GCN & 0.7973 &0.8345 & 0.0497& 105.2 \\
            ST-GCN &\textbf{0.8137} & 0.8641 & 0.0429 & 96.9 \\
            SSTKG &0.8095& \textbf{0.8692} & \textbf{0.04245} & \textbf{95.7} \\
        \end{tabular}
        \begin{tabular}{lcccc}
            \toprule
            & \multicolumn{4}{c}{TFNSW dataset: daily prediction} \\
            \cline{2-5}
            method & acc@10 & acc@15 & RMS & RSD \\
            \midrule
            SVR & 0.7914 &0.8215 &0.05047 & 90.1 \\
            LSTM & 0.8145 & 0.8374 & 0.4059 & 87.2\\
            GRU &0.8609 &0.9285 & 0.03867 & 63.7\\
            T-GCN &0.8745 &0.948 & 0.03641 & 67.5\\
            ST-GCN & 0.8991 & \textbf{0.9625} & 0.03583 & \textbf{52.8} \\
            SSTKG & \textbf{0.9051} & 0.9571 & \textbf{0.03488} & 54.3\\
            \bottomrule
        \end{tabular}
    \end{minipage}
\end{table*}

\subsection{Evaluation}

The accuracy rate for a prediction in the study is quantified using ACC\@n metric, which is defined as, if the predicted value falls within a specific range of the real value, it is deemed accurate. The range is determined by equation \ref{ACC}, allowing for a flexible yet rigorous assessment of prediction accuracy. Root Mean Square (RMS) and Relative Standard Deviation (RSD) are also used as supplementary evaluation metrics, defined in equation \ref{RMS} and \ref{RSD}.

\begin{equation} \label{ACC}
  r_{predict} \in (r_{real}(1-n\%), r_{real}(1+n\%))\\
\end{equation}

\begin{equation} \label{RMS}
  RMS = \sqrt{\frac{\Sigma_{i=1}^n(o_i-p_i)^2}{\Sigma_{i=1}^n(o_i)^2}}\\
\end{equation}

\begin{equation} \label{RSD}
  RSD = \sqrt{\frac{\Sigma_{i=1}^n(o_i-p_i)^2}{N}}\\
\end{equation}

To benchmark the performance of our model (SSTKG), several established models are used for comparison: (1) Support Vector Regression Machine(SVR). (2)Long Short-Term Memory (LSTM) network. (3) Gated Recurrent Unit(GRU) \cite{cho2014properties}. (4) TGCN \cite{zhao2019t}, the fusion of GCN and GRU. (5) STGCN \cite{yu2017spatio} which combines two TCNs and one GCN.

\subsection{Case study}

In order to validate the interpretability of the proposed model, a case study was conducted using the Spend-Ohio data in 2023-1. Specific stores served as exemplars. Following the knowledge graph construction and training of the influences and embeddings, the distance thresholds were adjusted to modify the quantity of entities deemed related in the knowledge graph. Three groups of entities near the center store are chosen, and prediction records after removing each of them are prepared and analyzed. By repeating the construction process with these variations, differences in outcomes aim to elucidate the model's explainability.

\section{Result}

\subsection{Experiment Results}

\subsubsection{Safegraph: Spend-Ohio dataset}

In Spend-Ohio dataset, the first 25 days are used to construct and train the SSTKG for monthly data, while the rest data is used for testing(which is 6 days, 3 days, and 6 days in the three subsets). To help compare and reduce the effect of null values, when calculating the RMS and RSD, the score's selling records is normalized to a range of \textbf{(0,20)}. The results are shown in Table~\ref{combined_spendOhio}.

\subsubsection{TFNSW dataset}

In TFNSW data, two separate experiments were done. The first one used the hourly data collected 24/7. 40 weeks' data were used to train, and then a 24-hour prediction in the following days was generated. In the second experiment, hourly records were added to daily ones, then the daily records were used to train. It is similar to the scale in the Spend-Ohio dataset. Similarly, the traffic volume was also normalized to \textbf{(0,20)}. The accuracy result (acc\@10 and acc\@15) and the RMS and RSD for normalized data are shown in Table~\ref{combined_TFNSW}.

\subsection{Result analysis}

Notably, from the results, the prediction of T-GCN, ST-GCN, and SSTKG are consistently outperformed the SVR, GRU and LSTM models. This is because these three models only focus on temporal record correlations while failing to consider spatial relations. SSTKG, as well as T-GCN and ST-GCN, model both spatial and temporal characteristics to ensure the data effectiveness, which are more suitable for datasets. SSTKG, in particular, demonstrates the ability to balance and integrate spatial-tempoal dimensions. Is outperformed T-GCN. Its performance is noteworthy, especially in acc@15 and RSD metrics on the Spend-Ohio dataset. It outshines others in acc@15, RMS, and RSD for the hourly predictions in the TFNSW dataset, and acc@10 and RMS metrics in daily predictions.

\subsection{Interpretability: case study}

\begin{figure}[htbp]   
\centering         
\includegraphics[width=0.4\textwidth]{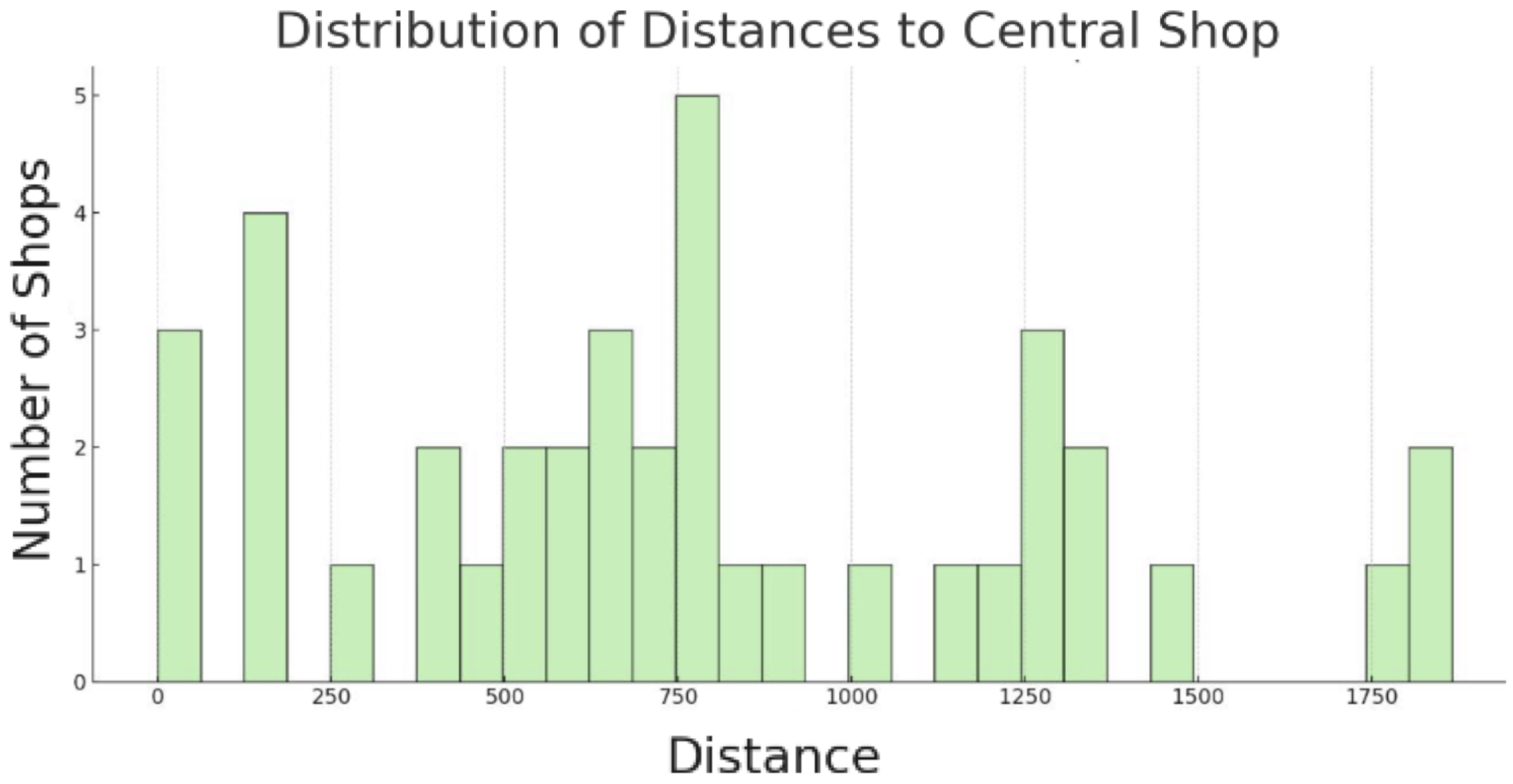}
\caption{Related entities' distances with sample shop}
\label{distances}
\end{figure}

\begin{figure}[htbp]   
\centering         
\includegraphics[width=0.4\textwidth]{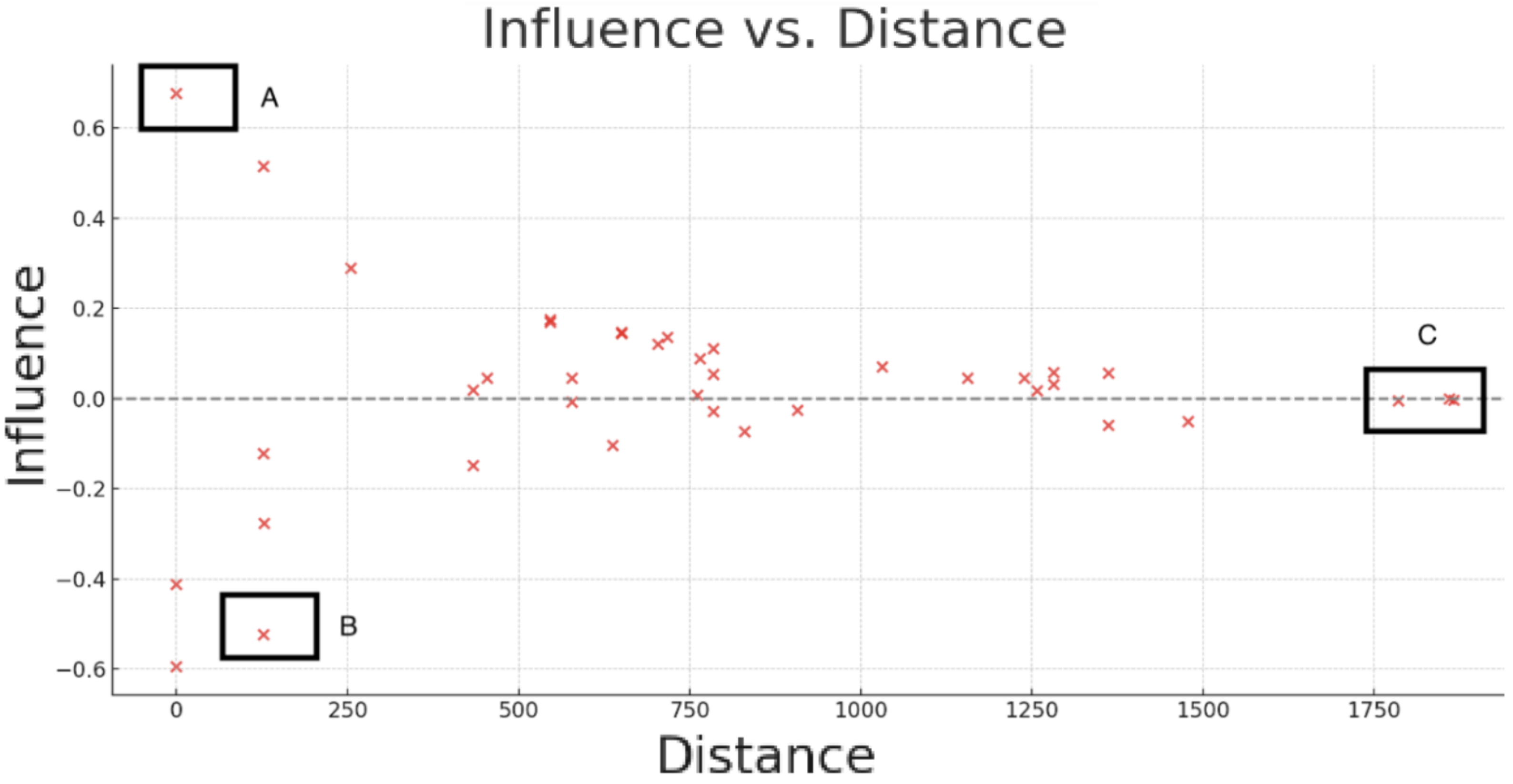}
\caption{Related entities' influence to sample shop}
\label{influences}
\end{figure}

To demonstrate the interpretability of the SSTKG model, this section presents a detailed case study involving a specific entity in the Spend-Ohio dataset. A full type service restaurant is selected carefully and set as the center entity (placekey: 225-222@63j-xxx-xxx; NAICS:722511). The sample is selected due to following reasons: \textbf{1. Completeness of records}: The selected stores as well as linked entities have complete records in all parts of Spend-Ohio dataset. \textbf{2. General category}: The selected sample belongs to full type service restaurant, which is a general type in the data. \textbf{3. Proper related entities}: The store located in a relatively popular area. Distances of nearby entities are shown in Figure~\ref{distances}. There are 36 entities in SSTKG that have influence with this shop. Figure~\ref{influences} shows the influence values that are calculated and extracted from SSTKG. 

\begin{table}[ht]
\centering
\caption{Real values vs Predicted and Adjusted Data}
\label{casestudy}
\begin{tabular}{cccccc}
\hline
\textbf{Day} & \textbf{Real values} & \textbf{$R_0$} & \textbf{$R_a$} & \textbf{$R_b$} & \textbf{$R_c$} \\
\hline
1 & 263.95 & 277.47 & 257.44 & 289.69 & 281.82 \\
2 & 495.81 & 530.09 & 517.74 & 539.26 & 528.27 \\
3 & 257.85 & 239.37 & 228.33 & 245.70 & 242.62 \\
4 & 352.82 & 372.83 & 352.14 & 381.54 & 373.38 \\
5 & 196.54 & 188.06 & 172.63 & 191.41 & 190.35 \\
6 & 409.67 & 435.99 & 413.76 & 443.57 & 434.63 \\
7 & 200.7 & 189.11 & 180.15 & 198.97 & 185.34 \\
\hline
\end{tabular}
\end{table}

From the above results, generally, entities close to the sample entity tend to have larger absolute influence values, whereas those entities located further away exhibit minimal or no influence on the sample. Three groups of entities are chosen for further analysis, marked as A, B and C. Entity A and Entity B are close to the center store, having high 'influence' values. Group C, although their temporal records are significant, their spatial and categorical attributes play a crucial role in the model's calculations, resulting in them having a minimal influence value. By integrating the above influences with trained embeddings, the sample's selling is predicted based on Equation~\eqref{eq5}, \eqref{eq6} and~\eqref{eq6-2} (first calculate embeddings then decode records). However, if some related entities, like A, B and group C were masked, the predicted result would change. Table~\ref{casestudy} shows the change of prediction after masking entities. While the former predicted data is $R_0$, predicted data after removing A, B and~C are $R_a,R_b \text{ and } R_c$.

A hypothesis test is set to show the difference between predicted data. The null hypothesis are: $H_{0a}: R_0 < R_a$; $\text{ }H_{0b}: R_0 > R_b$; $\text{ }H_{0c}: R_0 != R_c$, while alternative hypothesis are $H_{1a}: R_0 > R_a$; $\text{ }H_{1b}: R_0 < R_b$; $\text{ }H_{1c}: R_0 = R_c$. Table~\ref{t-test} shows the p-value after t-test under 95\% confidence level. For all three null hypotheses, the p-value of t-test is greater than 0.05, thus are all rejected, drawing the conclusion that, by masking entity A, the predicted value for sample's selling decreased($R_0 > R_a$), while by masking B the predicted value increased ($R_0 > R_b$) -- those who have positive influence on SSTKG would increase prediction, which means ``prosperity in one shop leads to prosperity to another'', and vice versa. On the other hand, in group C, where entities have small influence values, the prediction value changed a little after masking them (more than 95\% confidence to confirm that $R_0=R_c$).

\begin{table} [ht]
\caption{Result for t-test}
\label{t-test}
  \begin{tabular}{ccc}
    \toprule
    hypothesis & p-value & result\\
    \midrule
    $H_{0a}: R_0 < R_a$ & 0.9998975 & reject $H_{0a}$, accept $H_{1a}$\\
    $H_{0b}: R_0 > R_b$ & 0.999873 & reject $H_{0b}$, accept $H_{1b}$\\
    $H_{0c}: R_0 != R_c$ & 0.6717662 &reject $H_{0c}$, accept $H_{1c}$\\
  \bottomrule
\end{tabular}
\end{table}

\section{Conclusions and future work}

In this paper, a new knowledge graph framework is proposed, i.e., Simple Spatio-Temporal Knowledge Graph (SSTKG), which leverages 3 kinds of embeddings (static, temporal in and out embeddings) to model entities, as well as using ``influence'' to model the spatio-temporal relations between entities. A comprehensive evaluation using real-world data has underscored the efficacy of the proposed SSTKG in prediction tasks and highlighted its interpretability. 
Future endeavors will focus on (1)Refining the SSTKG construction algorithm. (2) Enhancing dynamism in SSTKG to reflect entities' exhibit temporal mobility such as user POI trajectories in which locations are shifting. (3)Balance between model size and efficiency.

\section{Ethical Use of Data}


\smallskip
The Spend-Ohio dataset from SafeGraph was utilized for this study. While it provides granular transaction data, all transactions and associated credit or debit card details have undergone rigorous anonymization to safeguard consumer privacy. Specific details about the merchants (like location and brand) within the Spend-Ohio dataset were masked from the study. All information regarding merchants and consumers was handled with strict confidentiality, ensuring that no privacy boundaries were breached. No credit information of merchants and consumers is involved in this paper.

Additionally, the TFNSW dataset used in the experiment is a publicly available dataset that contains neither personal nor private details. The dataset only incorporates generic traffic flow without identifiable details, without specific identifiable details such as license plate numbers or exact timestamps of certain car passes.




\begin{acks}
We acknowledge the support of Cisco Research Gift (CG\# 75677887), the Australian Research Council (ARC) Centre of Excellence for Automated Decision-Making and Society (ADM+S) (CE200100005), and the resources and services from the National Computational Infrastructure (NCI), which is supported by the Australian Government.
\end{acks}

\printbibliography
\end{document}